\documentclass{article}

\usepackage{booktabs}
\usepackage{siunitx}
\usepackage{microtype}
\usepackage{graphicx}
\usepackage{booktabs} 
\usepackage{amsmath}
\usepackage{amssymb}
\usepackage{caption}
\usepackage{subcaption}
 \usepackage{color}
\usepackage{hyperref}

\usepackage[accepted]{icml2019}

\graphicspath{ {./figures/} }

\icmltitlerunning{A Utility-Preserving GAN for Face Obscuration}

\begin{document}

\twocolumn[
\icmltitle{A Utility-Preserving GAN for Face Obscuration}



\icmlsetsymbol{equal}{*}

\begin{icmlauthorlist}
\icmlauthor{Hanxiang Hao}{viper}
\icmlauthor{David G\"{u}era}{viper}
\icmlauthor{Amy R. Reibman}{ece}
\icmlauthor{Edward J. Delp}{viper}
\end{icmlauthorlist}

\icmlaffiliation{viper}{Video and Image Processing Laboratory (VIPER), School of Electrical and Computer Engineering, Purdue University, West Lafayette, Indiana, USA}
\icmlaffiliation{ece}{School of Electrical and Computer Engineering, Purdue University, Indiana, USA}
\icmlcorrespondingauthor{Edward J. Delp}{ace@ecn.purdue.edu}

\icmlkeywords{Privacy Protection, Face Obscuration, Deep Learning, GAN}

\vskip 0.3in
]



\printAffiliationsAndNotice{}  

\begin{abstract}
From TV news to Google StreetView, face obscuration has been used for privacy protection. 
Due to recent advances in the field of deep learning, obscuration methods such as Gaussian blurring and pixelation are not guaranteed to conceal identity.
In this paper, we propose a utility-preserving generative model, UP-GAN, that is able to provide an effective face obscuration, while preserving facial utility. 
By utility-preserving we mean preserving facial features that do not reveal identity, such as age, gender, skin tone, pose, and expression. 
We show that the proposed method achieves the best performance in terms of obscuration and utility preservation. 
\end{abstract}

\section{Introduction} 
\label{Introduction}
Major developments in the machine learning field have uncovered severe flaws in current face obscuration approaches.
As shown by~\cite{McPherson_2016}, machine learning methods are able to defeat Gaussian blurring or pixelation based obscuration methods.
These obscuration techniques have been widely used by Internet news outlets, social media platforms, and government agencies.
An extreme resort to prevent information leaking is to simply gray out the entire facial region by setting all pixels in the facial area to a fixed value. 
However, this approach is rarely used because its visual effect is unpleasant, especially if there are many faces to be redacted.
Besides the identifiable information, facial images also contain information that does not reveal identity, such as age, gender, and skin tone. 
Often, we want to preserve these features in many applications involving visual understanding and data mining \cite{Du_2014}.

\begin{figure}[t]
    \vskip 0.1in
    \centering
    \includegraphics[width=1\linewidth]{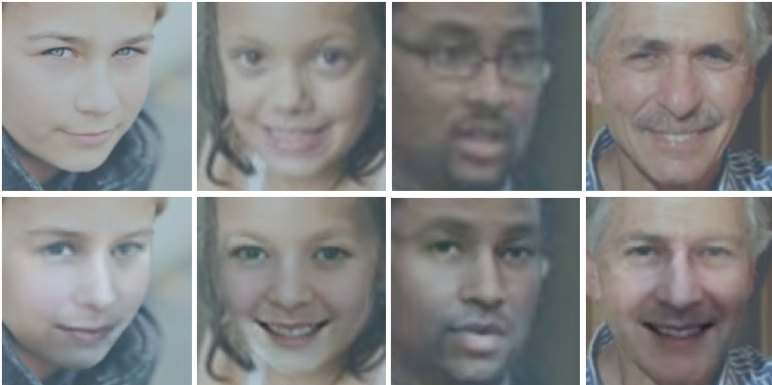}
    \caption{Obscuration effect of the proposed method. First row: original faces; second row: obscured faces.}
    \label{fig:intro}
    \vskip -0.2in 
\end{figure}

New obscuration methods are needed to remove identifiable facial information, while preserving the features that do not convey identity.
The proposed method, utility-preserving GAN (UP-GAN), aims to provide an effective obscuration by generating faces that only depend on the non-identifiable facial features. 
In this work, we define utility as the facial properties such as age, gender, skin tone, pose, and expression. 
We choose these properties because in practice, when dealing with a large number of identities, knowing these properties from the obscured images cannot reveal identity. 
One can also choose other properties to retain for different applications. 
As shown in Figure \ref{fig:intro}, UP-GAN is able to obscure the original faces by replacing them with synthetic faces that have the same utility. 

The main contributions of this work are summarized as follows. 
First, we develop a model that can produce new faces for obscuration. 
Then, we provide an evaluation to show the effectiveness of the proposed method with different training objectives. 
Finally, compared to different methods, we show that UP-GAN can provide the best performance in terms of face obscuration and utility preservation.

\section{Related Work} \label{Related Work}
Standard approaches, such as pixelation and Gaussian blurring, achieve good obscuration performance in terms of human perception. 
However, \cite{McPherson_2016} proposed a deep learning method with a simple structure that is able to defeat these obscuration techniques.
To provide better obscuration performance, a variety of approaches have been proposed to balance the need to remove identifiable information while preserving utility information. 

\subsection{Face De-identification}
\textbf{$k$-same Methods.} 
This family of approaches first groups faces into clusters based on non-identifiable information such as expression, and then generates a surrogate face for each cluster. 
These methods can guarantee that any face recognition system cannot do better than $1/k$ in recognizing who a particular image corresponds to \cite{Gross_2005}, where $k$ is the minimum number of faces among all clusters.
This property is also known as $k$-anonymity \cite{Samarati_1998}.  
In \cite{Newton_2005} and \cite{Gross_2005}, they simply compute the average face for each cluster. 
Therefore, their obscured faces are blurry and cannot handle various facial poses. 
In \cite{Du_2014}, the use of an active appearance model \cite{Cootes_2001} to generate more realistic surrogate faces is presented. 
A generative neural network, $k$-same-net, that directly generates faces based on the cluster attributes is described in \cite{Meden_2018}. 
These two methods are able to produce more realistic obscured faces with the property of $k$-anonymity, but cannot handle different poses.  

\textbf{GAN Methods.} 
Generative adversarial network (GAN) \cite{Goodfellow_2014} methods can provide more realistic faces. 
Their discriminator is designed to guide the generator by distinguishing real faces from generated faces. 
In \cite{Wu_2019}, a model that produces obscured faces directly from original faces based on conditional-GAN \cite{Mirza_2014} is proposed. 
They use a contrastive loss to enforce the obscured face to be different than the input face. 
However, since they need to directly input the original faces, the obscuration performance is not guaranteed. 
\cite{Sun_2018} present a two-stage model that is able to generate an obscured face without the original identifiable facial information, which prevents the leakage of identifiable information directly from faces.  
GANs have also been used for face manipulation in videos. 
These techniques aim to create believable face swaps without tampering traces, by altering age \cite{Antipov_2017} or skin color \cite{Lu_2018}. 
To prevent scenarios where these videos are used to create political distress or fake terrorism events, \cite{Guera_2018} design a deep learning model that is able to detect the altered frames using both the spatial and temporal information. 

Our proposed method tries to leverage the advantages of both types of methods. 
To achieve $k$-anonymity, it is designed to generate faces that depend only on the utility information without directly accessing original faces. 
Since it is also a GAN based method, with the discriminator guidance, it is able to produce more realistic faces than the $k$-same methods. 

\section{Proposed Method} \label{Method}
Recall that, in this implementation, we choose age, gender, skin tone, pose, and expression as the utility to be preserved. 
To better formulate our problem, we further divide the utility into two parts: attributes and landmarks. 
Attributes define the static part of the utility information that does not change with facial movement. 
Landmarks define a set of points of interest that describe the facial pose and expression. 
In order to obtain obscured faces, we first use an auxiliary system to detect the utility information: attribute vector $v_a$ and landmark vector $v_l$ from the original face $I_{real}$.
Since, in this work, we are not focusing on this auxiliary system, we use the UTKFace dataset \cite{Zhang_2017} which provides the needed attributes (age, gender, and skin tone) and landmarks (7 points) to train and test our model. 
The fake face $I_{fake}$ is then generated by the UP-GAN model using the attribute and landmark vectors.  
Given that the generated face has the same pose and expression, we can swap it with the original face to perform de-identification using face swapping algorithms \cite{Perez_2003, Bitouk_2008, Korshunova_2017}. 
Figure \ref{fig:intro} shows the swapping results using \cite{Perez_2003}. 

\begin{figure}[t]
    \vskip 0.2in
    \begin{center}
    \centerline{\includegraphics[width=0.9\columnwidth]{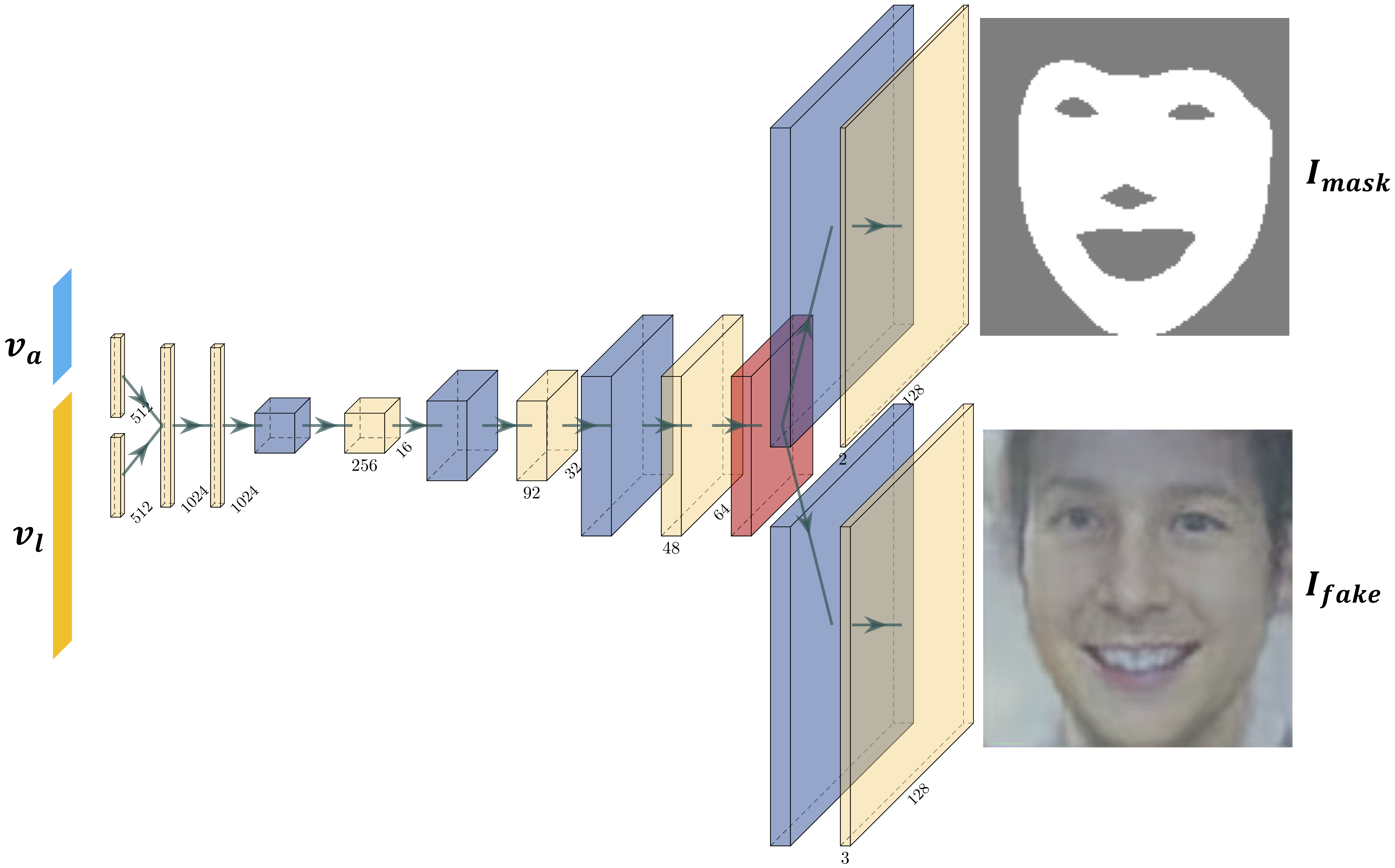}}
    \caption{Generator architecture of the UP-GAN model. Yellow vectors indicate the activation of fully-connected layers. Blue blocks indicate the activation from de-convolution layers (upsampling + convolution). Yellow blocks show the activation from following convolution layers after the de-convolution layer. The red block shows the output from the max pooling layer.}
    \label{fig:generator}
    \end{center}
    \vskip -0.3in
\end{figure}

Figure \ref{fig:generator} shows the generator architecture of the UP-GAN model, which is based on the architecture proposed by \cite{Dosovitskiy_2017}. 
Similar to the previous work, UP-GAN jointly learns the fake face and its binary mask. 
However, we modify the structure of the fully-connected layers to input the attribute and landmark vectors. 
As suggested by \cite{Meden_2018}, we also add a max pooling layer with stride 1 for dimension reduction before generating the output image and mask. 
More specifically, we first use two fully-connected layers to encode the input vectors and then apply de-convolution, followed by another convolution layer to upsample the feature maps.
The de-convolution layer contains an upsampling layer with stride 2 and a convolution layer with a kernel size of 5. 
For the following convolution layer after the de-convolution layer, we choose the kernel size to be 3. 
Note that the final output size of the generated face is $128\times128\times3$ and the size of the binary mask is $128\times128\times2$. 

The loss functions for the generator $G$ and discriminator $D$ are defined as:
\begin{align*}
    & \mathcal{L}_G = \mathbb{E}_{v_a, v_l}[\log{D(G(v_a, v_l)))}]+ \\
    & \ \ \ \ \ \ \ \ \ \ \ \ \ \ \ \ \ \ \ \ \ \ \ \ \ \  \lambda_1\mathcal{L}_{2} + \lambda_2\mathcal{L}_{M} + \lambda_3\mathcal{L}_{P}, \\
    & \mathcal{L}_D = \mathbb{E}_{I_{real}}[\log{D(I_{real})}] + \\
    & \ \ \ \ \ \ \ \ \ \ \ \ \ \ \ \ \ \ \ \ \ \ \ \ \ \  \mathbb{E}_{v_a, v_l}[\log{(1-D(G(v_a, v_l)))}],  \\ 
\end{align*}
\vspace{-0.25in}
\newpage 
where 
\begin{align*}
    & \mathcal{L}_{2} = \Vert I_{real} - I_{fake} \Vert_2^2, \\
    & \mathcal{L}_{M} = -\frac{1}{N} \sum_{i=1}^{N} y_i\log{(p_i)} + (1 - y_i)\log{(1 - p_i)}. \\
    & \mathcal{L}_{P} = \sum_{l \in \Omega} \Vert \phi_l(I_{fake}) - \phi_l(I_{real}) \Vert_2^2, \\
\end{align*}
\vskip -0.3in
$\mathcal{L}_{2}$ is the reconstruction loss for learning the image content.  
$\mathcal{L}_{M}$ is the binary cross entropy loss for learning the facial mask where $p_i$ is the predicted probability of the $i$-th pixel in the binary mask, $y_i$ is the ground truth label, and $N$ is the total number of pixels. 
$\mathcal{L}_{P}$ is the perceptual loss for learning the facial details, where $\Omega$ is a collection of convolution layers from the perceptual network and $\phi_l$ is the activation from the $l$-th layer.  
The perceptual loss was originally proposed by~\cite{Johnson_2016} for learning high level features extracted from a network pretrained on the ImageNet dataset~\cite{Russakovsky_2015}. 
In our work, the perceptual network is pretrained on a face identification dataset to enforce that the generated face contains similar facial features to the original face. 
More specifically, we choose the pretrained VGG-19 network~\cite{Simonyan_2015} and finetune it with the FaceScrub dataset~\cite{Ng_2014}. 
Lastly, $\lambda_1$, $\lambda_2$, and $\lambda_3$ are the scalar weights for their corresponding losses. 
Note that in our implementation, we have chosen $\lambda_1=5$, $\lambda_2=1$, and $\lambda_3=1$ to ensure that the terms in $\mathcal{L}_G$ are within the same numerical order of magnitude.

\section{Experiments} \label{Experiment}
In this section, we will evaluate different loss functions and analyze the obscuration performance of the generated faces compared to Gaussian blurring, pixelation, $k$-same method and $k$-same-net method. 

\subsection{Datasets}
The UTKFace dataset~\cite{Zhang_2017} contains 23,708 images with annotations of 68-point facial landmarks and attributes of age, gender, and skin tone. 
To obscure the identifiable information present in the facial landmarks, we reduce the input landmark points from 68 points to 7 points.
These include the centers of the eyes, the center of the nose, and four points around the mouth.
Therefore, the dimensionality of the attribute vector is 3 and of the landmark vector is 14. 
From the perspective of $k$-anonymity, reducing landmark points is similar to increasing $k$.
When we increase $k$, the size of each cluster also increases, since they are grouped based on attribute and landmark vectors. 
Therefore, the upper bound of identification rate ($1/k$) decreases, meaning that the obscuration performance improves. 

To verify the obscuration performance, we use the FaceScrub dataset for face identification.
Note that this dataset contains 106,806 images from 530 identities. 
As this dataset does not provide attributes and landmarks, we use fixed attribute values and detect facial landmarks using the \textit{Dlib} toolkit \cite{King_2009}. 
We can produce fake faces using the fixed attributes and detected landmarks. 
We then use a face identification model (VGG-19) to determine if we are able to identify these generated faces. 

\subsection{Data Augmentation}
\begin{figure}[t]
    \vskip 0.1in
    \begin{center}
    \centerline{\includegraphics[width=0.5\columnwidth]{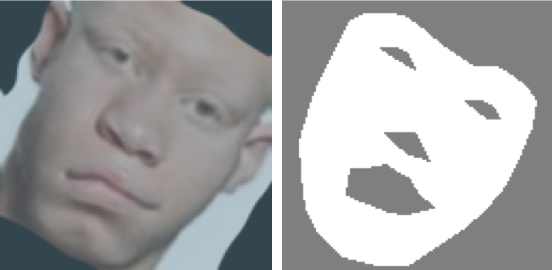}}
    \caption{Example of the augmented face with elastic distortion and random rotation (left) and its binary mask (right).}
    \label{fig:augmentation}
    \end{center}
    \vskip -0.3in
\end{figure}

To prevent UP-GAN from simply memorizing the original face and replicating the output face using the input vectors, we use data augmentation on the original image $I_{real}$ to increase its variation. 
First, we use elastic distortion~\cite{Simard_2003} to add variety to the facial landmarks. 
As shown in Figure~\ref{fig:augmentation}, the wave-like structure distorts the landmark points (e.g.\ the shape of the mouth). 
We also add random rotations, ranging from \ang{-30} to \ang{30}, to increase the variation of facial poses. 

\subsection{Results and Discussion}
\begin{figure}[t]
    \vskip 0.1in
    \begin{center}
    \centerline{\includegraphics[width=0.85\columnwidth]{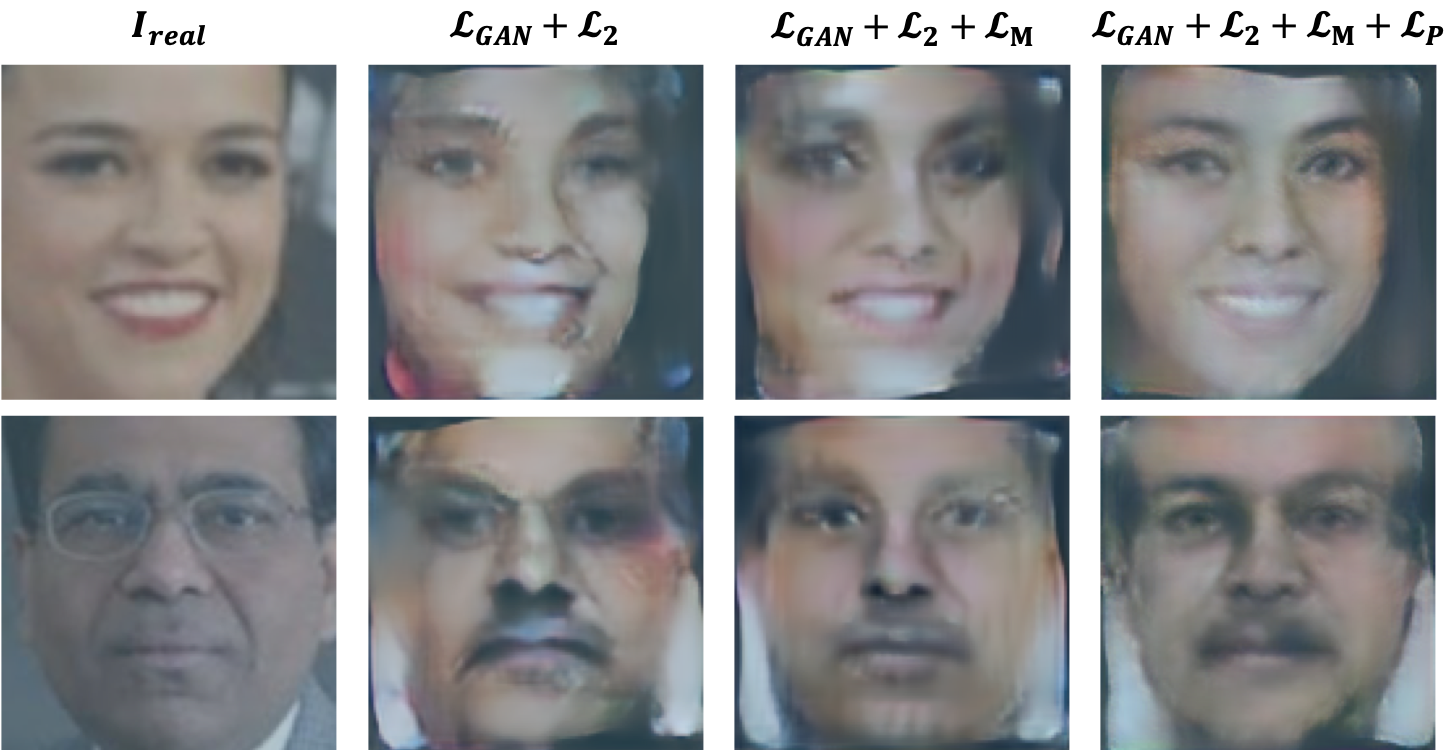}}
    \caption{Generated faces with different loss functions.}
    \label{fig:loss_compare}
    \end{center}
    \vskip -0.3in
\end{figure}

In Figure \ref{fig:loss_compare}, we compare the results using different loss functions to show the effectiveness of training with the perceptual network and binary mask. 
We can also see that, compared to the original face, the generated face with adversarial loss and $L_2$ reconstruction loss can preserve the facial utility.
However, the facial details such as the outlines are partially missing.
By adding the mask loss, we can enhance the facial boundary, like the cheek and chin. 
If we add the perceptual loss, the generated face visually looks more realistic with fewer ripple-like artifacts. 

\begin{figure}[t]
    \vskip 0.1in
    \begin{center}
    \centerline{\includegraphics[width=0.9\columnwidth]{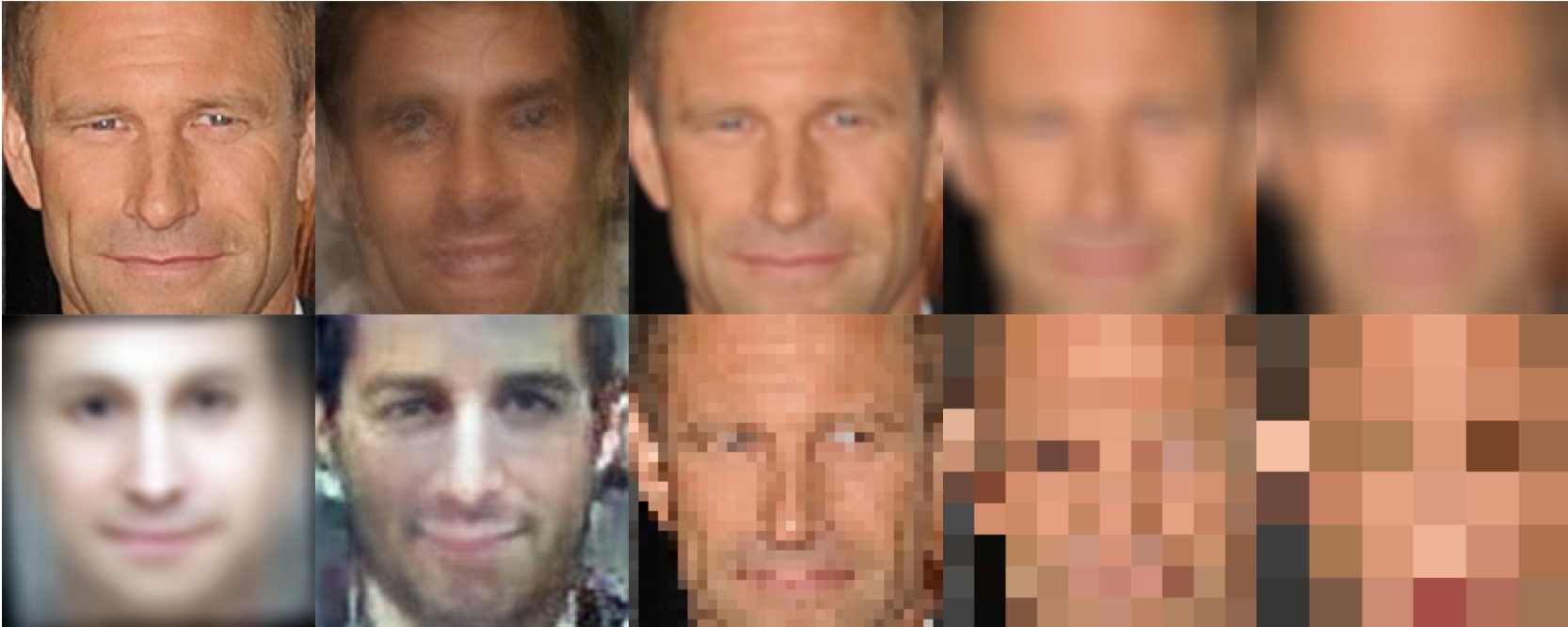}}
    \caption{Examples of obscured faces. Top row: original image, $k$-same ($k=10$) and Gaussian blurring (kernel sizes: 5, 15 and 25). Bottom row: $k$-same-net, UP-GAN and pixelation (pixel sizes: 5, 15 and 25).}
    \label{fig:obscuration}
    \end{center}
    \vskip -0.3in
\end{figure}

We also evaluate the obscuration performance to see how well UP-GAN can conceal the original faces. 
We consider two threat models:
I) the attacker (identifier) has no information about the obscuration methods and
II) the attacker knows the obscuration methods. 
In threat model I, we train the identifier on the pristine images and test it on the obscured faces. 
In threat model II, we train and test the identifier on both clear and obscured images. 
To provide a fair comparison with the other obscuration methods, we use the generated faces $I_{fake}$ as the obscured images, but we do not swap them into the original images. 
This is because the unobscured area (non-facial region) may contain identifiable information. 
Figure~\ref{fig:obscuration} shows the visual quality of the obscured images with different methods including Gaussian blurring, pixelation, $k$-same method~\cite{Gross_2005}, $k$-same-net method~\cite{Meden_2018}s, and UP-GAN.
We modify the input layers of the $k$-same net method to input the same attribute and landmark vectors as UP-GAN. 
The obscured face from $k$-same method is really blurry like the areas of eyes, although the skin tone is preserved. 
The result from $k$-same-net method contains more facial structures, but compared to UP-GAN, the facial boundary is not clear.
To further quantify the visual performance, we compute the Fr\'{e}chet inception distances (FID)~\cite{Heusel_2017} of the obscured faces. 
With the assumption that the real and obscured faces are two sets of realizations coming from two distributions, FID measures the distance of these two distributions. 
Therefore, we can use FID to estimate how realistic the obscured faces are.  
As shown in Table~\ref{table:threat_model}, UP-GAN achieves the minimum FID value, which confirms that the obscured face has the best visual quality. 

Table~\ref{table:threat_model} also compares the obscuration performance of UP-GAN against other methods. 
For the threat model I, Gaussian blurring with kernel size 5 and 15 fail to provide an effective obscuration, while all other methods achieve good performance. 
For the threat model II, the obscuration performance degrades for all methods, while pixelation with pixel size 25, $k$-same, $k$-same-net, and UP-GAN still achieve relatively good results. 
However, as shown in Figure \ref{fig:obscuration}, for pixelation-25 there are only $5\times5$ blocks representing the facial region. 
As with $k$-same and $k$-same-net, the visual quality of pixelation-25 is worse than UP-GAN. 

\begin{table}[t]
    \centering
    \begin{tabular}{@{}cccc@{}}
    \toprule
    Method        & Threat Model I& Threat Model II & FID \\ \midrule
    None          & 0.955         & 0.955           &  - \\
    Gaussian-5    & 0.914         & 0.979           & 212.36 \\
    Gaussian-15   & 0.360         & 0.983           & 386.31 \\
    Gaussian-25   & 0.046         & 0.923           & 358.86 \\
    Pixelation-5  & 0.010         & 0.897           & 154.26 \\
    Pixelation-15 & 0.003         & 0.694           & 576.28 \\
    Pixelation-25 & 0.003         & 0.191           & 486.41 \\
    $k$-same      & 0.003         & 0.028           & 91.41  \\
    $k$-same-net  & 0.003         & 0.238           & 252.90  \\
    UP-GAN        & 0.004         & 0.245           & 68.78  \\ \bottomrule
    \end{tabular}
    \caption{Face identification accuracy and FID of the obscured faces for different obscuration methods. Note that the method ``None'' means no obscuration and $k=10$ for the $k$-same method.}
    \label{table:threat_model}
\end{table}

\section{Conclusion} \label{Conclusion}
Gaussian blurring or pixelation cannot guarantee obscuration and preserve utility such as age, gender, skin tone, pose, and expression.
Our proposed approach, UP-GAN, is able to generate faces that preserve utility while also removing identifiable information from the original faces. 
By swapping the generated face back on the original image, we can produce an effective obscuration that not only removes personal identifiable information, but also retains the information that does not reveal identity. 
Based on our results, we show that UP-GAN is able to achieve a better performance than Gaussian blurring, pixelation, and $k$-same method in terms of face obscuration and utility preservation.  

\section*{Acknowledgments} 
This material is based on research sponsored by the Department of Homeland Security (DHS) under agreement number 70RSAT18FR0000161. The U.S. Government is authorized to reproduce and distribute reprints for Governmental purposes notwithstanding any copyright notation thereon. The views and conclusions contained herein are those of the authors and should not be interpreted as necessarily representing the official policies or endorsements, either expressed or implied, of DHS or the U.S. Government. 

\bibliography{paper}
\bibliographystyle{icml2019}

\end{document}